\pdfoutput=1
\documentclass{article}
\usepackage[numbers]{natbib}

\usepackage[final]{neurips_2021}

\usepackage{microtype}      
\usepackage{graphicx}
\usepackage{hyperref}
\usepackage{url}
\usepackage{xcolor}

\title{StyleCLIPDraw: Coupling Content and Style in Text-to-Drawing Synthesis}

\author{%
  Peter Schaldenbrand \\
  Robotics Institute \\
  Carnegie Mellon University\\
  Pittsburgh, PA 15213 \\
  \texttt{pschalde@andrew.cmu.edu} \\
   \And
  Zhixuan Liu \\
  School of Data Science \\
  The Chinese University \\ of Hong Kong \\
  Shenzhen, China \\
  \texttt{zhixuanliu@cuhk.edu.cn} \\
  \And
  Jean Oh \\
  Robotics Institute \\
  Carnegie Mellon University\\
  Pittsburgh, PA 15213 \\
  \texttt{jeanoh@nrec.ri.cmu.edu} \\
}

\begin{document}
\maketitle \vspace{-35pt}%
\section*{
\centering
\color{red}
Please refer to our newer, more comprehensive manuscript of StyleCLIPDraw available at \url{https://arxiv.org/abs/2202.12362}
}

\begin{abstract}
  Generating images that fit a given text description using machine learning has improved greatly with the release of technologies such as the CLIP\cite{radford2021-clip} image-text encoder model; however, current methods lack artistic control of the style of image to be generated.  We introduce StyleCLIPDraw which adds a style loss to the CLIPDraw\cite{frans2021-clipdraw} text-to-drawing synthesis model to allow artistic control of the synthesized drawings in addition to control of the content via text. 
  Whereas performing decoupled style transfer on a generated image only affects the texture, our proposed coupled approach is able to capture  a  style in both texture and shape, suggesting that the style of the drawing is coupled with the drawing process itself.
  More results and our code are available at https://github.com/pschaldenbrand/StyleCLIPDraw

\end{abstract}

 \vspace{-15pt}%
\begin{figure}[hb]
    \centering
    \includegraphics[width=12cm]{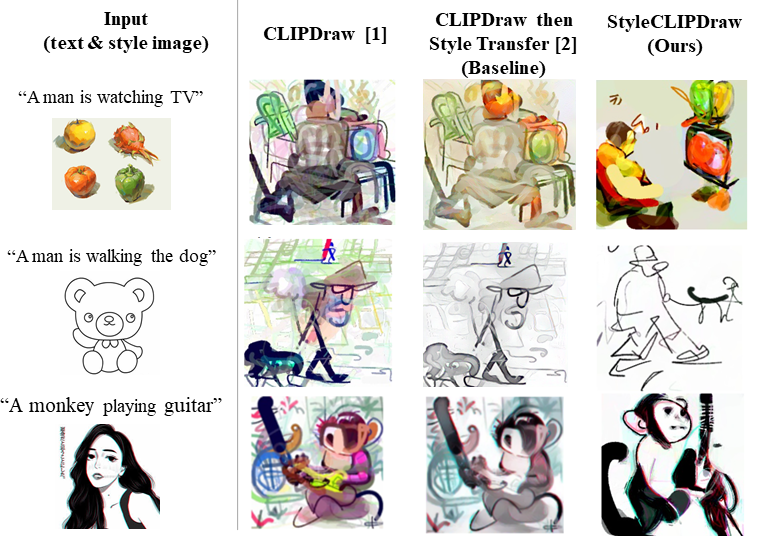}
    \caption{Comparing styled text-to-drawing results. The baseline is formed by using CLIPDraw \protect{ \cite{frans2021-clipdraw}} to convert the input text into an image, then performing style transfer \protect{\cite{kolkin2019-strots}}. Our model, StyleCLIPDraw, couples style and content by generating the drawing using both the text and style simultaneously.
    }
    \label{fig:style_transfer_results}
\end{figure}

\textbf{Introduction}
Text-to-drawing synthesis models provide a method for anyone to generate artistic visuals by describing the contents in natural language.  While drawing by hand or using software to render images can require specialized skills and training that not everyone may have the means for, text-to-drawing synthesis can serve as a method for individuals to generate visuals by proxy of a skill that they do have, language. Models such as CLIPDraw \cite{frans2021-clipdraw} or Dall-E \cite{ramesh2021-dalle} can create images based on text inputs, but these existing approaches lack additional control of the generation process such as style.  We introduce an additional input to the CLIPDraw text-to-drawing synthesis model to alter the style of the generated visuals.

\begin{figure}
    \centering
    \includegraphics[width=\textwidth]{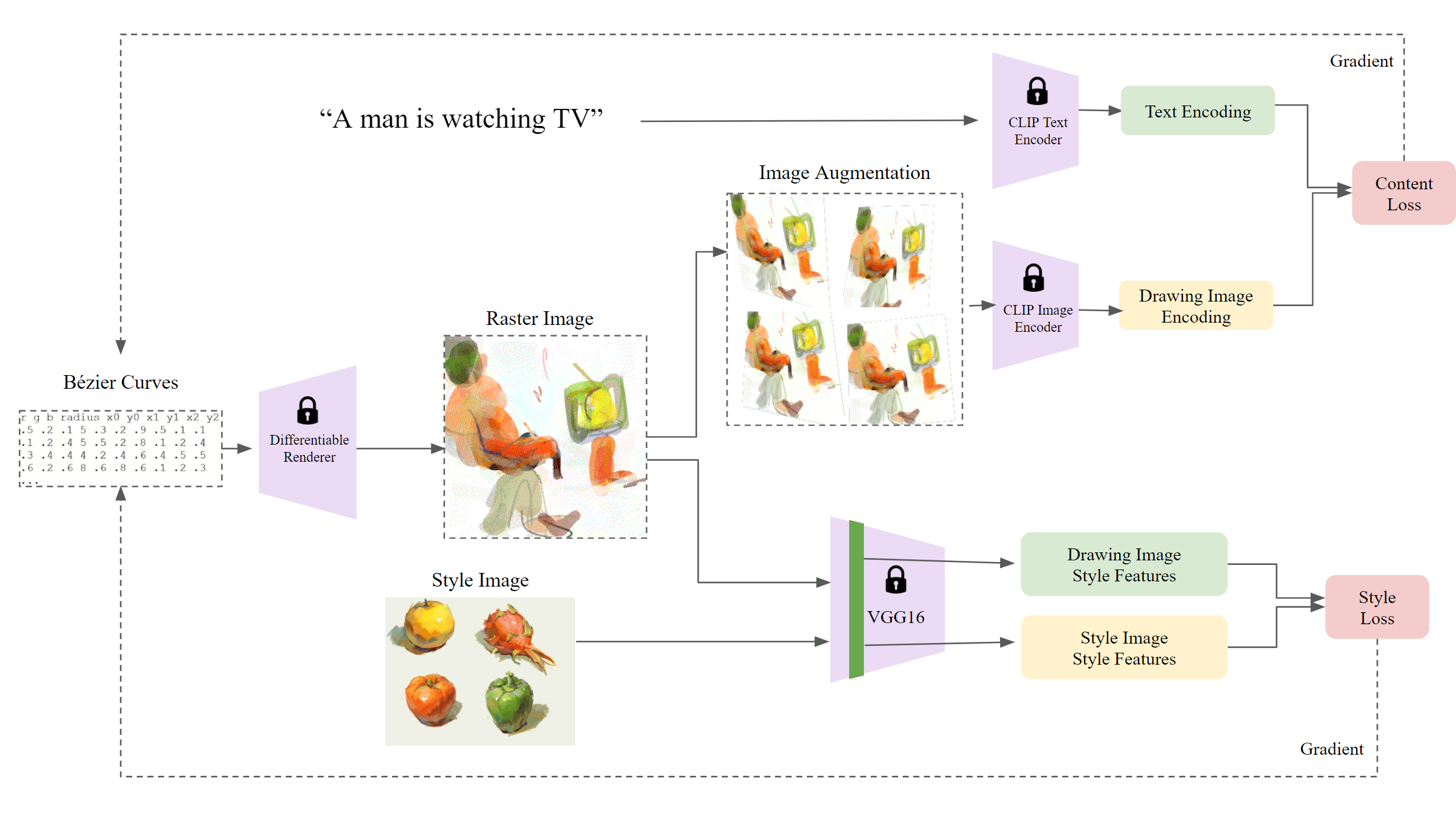}
    \caption{The StyleCLIPDraw model architecture. The B\'ezier curve representation is converted into a raster image and is then used to compute two losses: One loss for aligning the content of the image with the user's text prompt and the other for aligning style.}
    \label{fig:model}
\end{figure}

\textbf{Method}
Unlike most other image generation models, CLIPDraw produces drawings consisting of a series of B\'ezier curves defined by a list of coordinates, a color, and an opacity.  The drawing begins as randomized B\'ezier curves on a canvas and is optimized to fit the given style and text. The StyleCLIPDraw model architecture is shown in Fig. \ref{fig:model}.  The brush strokes are rendered into a raster image via  differentiable model \cite{li2020-renderer}.  There are two losses for StyleCLIPDraw that correspond to each input.  The text input and the augmented raster drawing are fed the the CLIP model and the difference in embeddings are compared using cosine distance to compute a loss that encourages the drawing to fit the text input.  The image is augmented to avoid finding shallow solutions to optimizing through the CLIP model.  The raster image and the style image are fed through early layers of the VGG-16 \cite{simonyan2014-vgg16} model (per the STROTSS style-transfer algorithm \cite{kolkin2019-strots}) and the difference in extracted features form the loss that encourages the drawings to fit the style of the style image.

\begin{figure}
    \centering
    \includegraphics[width=\textwidth]{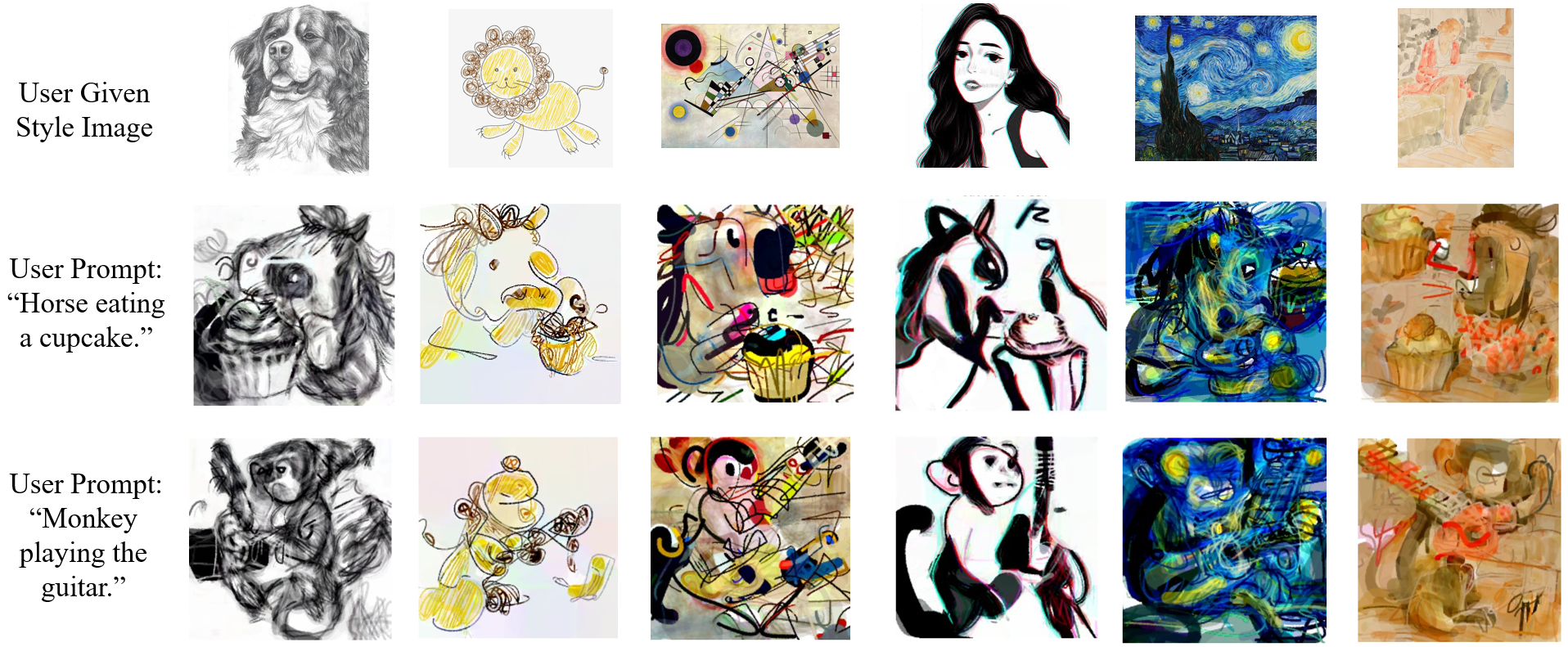}
    \caption{The left column specifies two user given prompts and the top row depicts style images.  StyleCLIPDraw generates the images to contain the content of the language input and the style of the style image.}
    \label{fig:additional_results}
\end{figure}

\textbf{Results}
Because CLIP\cite{radford2021-clip} is trained on such a large, diverse dataset, StyleCLIPDraw works on almost arbitrarily general text prompts. While the StyleCLIPDraw results shown in Fig. \ref{fig:style_transfer_results} and Fig. \ref{fig:additional_results} lack some realism with respect to fitting the text prompt exactly, all images show strong evidence of the given style and have a resemblance to the text which describes them.  The content of the StyleCLIPDraw drawings is no less recognizable than that of the CLIPDraw drawings, indicating that adding the style loss did not adversely affect the optimization for the text description loss.  

\textbf{Conclusions}
In the second example of Fig. \ref{fig:style_transfer_results}, the baseline results in a gray drawing with shading, but the given style image was a black, line drawing.  
StyleCLIPDraw was able to synthesize a proper line drawing that fit the content from the input text description.  
Performing style transfer on the CLIPDraw drawings changes the colors and textures of the drawings but does not change the shape of the content which is required for drawings in a certain style.  Simultaneously optimizing style and content in the drawing synthesis process is able to properly affect the shape of the image to fit the given style, indicating that the style of the drawing is coupled with the drawing process itself.

\section*{Ethical Considerations}

StyleCLIPDraw relies heavily on the feedback from the CLIP\cite{radford2021-clip} image-text encoding model.  CLIP was trained on 400 million image-text pairs scraped from the internet, and this dataset is not made publicly available.  As pointed out in the original CLIPDraw paper\cite{frans2021-clipdraw}, the biases in this data will be reflected in the generated images from the model.  The biases of the CLIP model have been investigated\cite{radford2021-clip}, and it is important to recognize the presence of them when utilizing StyleCLIPDraw.

\bibliographystyle{plain}
\bibliography{ref.bib}

\end{document}